\documentclass{article} 
\usepackage{iclr2021_conferenceMy,times}

\usepackage{amsmath,amsfonts,bm}

\def\eqref#1{equation~\ref{#1}}

\def\1{\bm{1}}

\DeclareMathAlphabet{\mathsfit}{\encodingdefault}{\sfdefault}{m}{sl}
\SetMathAlphabet{\mathsfit}{bold}{\encodingdefault}{\sfdefault}{bx}{n}

\usepackage{hyperref}
\usepackage{url}
\usepackage{graphicx} 
\usepackage{xcolor} 
\usepackage{graphbox} 
\usepackage{comment} 

\title{Why Convolutional Networks Learn \\ Oriented Bandpass Filters: \\Theory and Empirical Support}

\author{Isma Hadji \thanks{The work in this article was done while Isma Hadji was a PhD student at York University. The views expressed (or the conclusions reached) are her own and do not necessarily represent the views of Samsung Research America, Inc.}\\
Samsung AI Center (SAIC) \\
Toronto, Ontario, Canada \\
\texttt{isma.hadji@samsung.com} \\
\AND
Richard P. Wildes \\
Department of Electrical Engineering and Computer Science \\
Centre for Vision Research \\
York University \\
Toronto, Ontario, Canada \\
\texttt{wildes@cse.yorku.ca} \\
}

\newcommand*{\vcenteredhbox}[1]{\begingroup
	\setbox0=\hbox{#1}\parbox{\wd0}{\box0}\endgroup}
\iclrfinalcopy 
\begin{document}

\maketitle

\begin{abstract}
It has been repeatedly observed that convolutional architectures when applied to image understanding tasks learn oriented bandpass filters. A standard explanation of this result is that these filters reflect the structure of the images that they have been exposed to during training: Natural images typically are locally composed of oriented contours at various scales and oriented bandpass filters are matched to such structure. We offer an alternative explanation based not on the structure of images, but rather on the structure of convolutional architectures. In particular, complex exponentials are the eigenfunctions of convolution. These eigenfunctions are defined globally; however, convolutional architectures operate locally.  To enforce locality, one can apply a windowing function to the eigenfunctions, which leads to oriented bandpass filters as the natural operators to be learned with convolutional architectures. From a representational point of view, these filters allow for a local systematic way to characterize and operate on an image or other signal. We offer empirical support for the hypothesis that convolutional networks learn such filters at all of their convolutional layers. While previous research has shown evidence of filters having oriented bandpass characteristics at early layers, ours appears to be the first study to document the predominance of such filter characteristics at all layers. Previous studies have missed this observation because they have concentrated on the cumulative compositional effects of filtering across layers, while we examine the filter characteristics that are present at each layer. 
\end{abstract}

\section{Introduction}
\label{sec:intro}
\subsection{Motivation}
\label{sec:motivation}
Convolutional networks (ConvNets) in conjunction with deep learning have shown state-of-the-art performance in application to computer vision, ranging across both classification (e.g., \cite{Krizhevsky2012,Tran2015,Ge2019}) and regression (e.g., \cite{Szegedy2013,Eigen2015,Zhou2017}) tasks. However, understanding of how these systems achieve their remarkable results lags behind their performance. This state of affairs is unsatisfying not only from a scientific point of view, but also from an applications point of view. As these systems move beyond the lab into real-world applications better theoretical understanding can help establish performance bounds and increase confidence in deployment.

Visualization studies of filters that have been learned during training  have been one of the key tools marshalled to lend insight into the internal representations maintained by ConvNets in application to computer vision, e.g., \cite{Zeiler2014,yosinski2015,MahendranV15,SHANG2016,Feichtenhofer2018}. Here, an interesting repeated observation is that early layers in the studied networks tend to learn oriented bandpass filters, both in two image spatial dimenstions, $(x,y)^{\top}$, in application to single image analysis as well as in three spatiotemporal dimensions, $(x,y,t)^{\top}$, in application to video. An example is shown in Fig.~1. 
Emergence of such filters seems reasonable, because local orientation captures the first-order correlation structure of the data, which provides a reasonable building block for inferring more complex structure (e.g., local measurements of oriented structure can be assembled into intersections to capture corner structure, etc.). Notably, however, more rigorous analyses of exactly why oriented bandpass filters might be learned has been limited. This state of affairs motivates the current paper in its argument that the analytic structure of ConvNets constrains them to learn oriented bandpass filters.

\begin{figure}[t] \label{fig:oriented}
	\centering
	\includegraphics[width=0.4\linewidth]{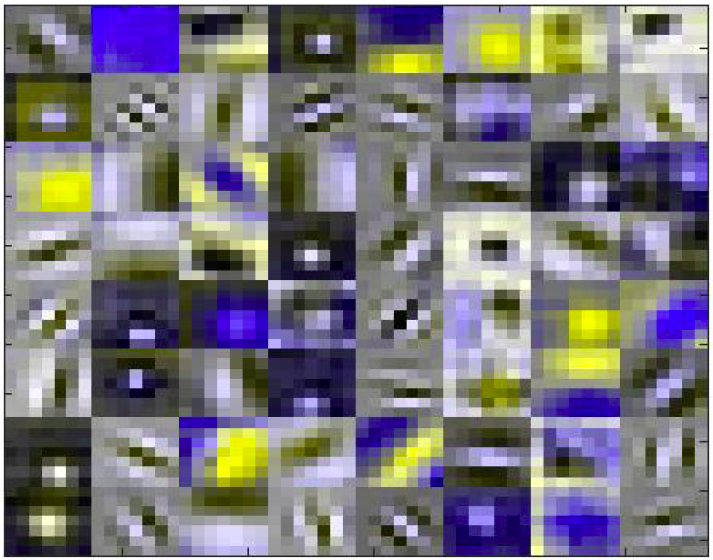}
	\caption{Visualization of pointspread functions (convolutional kernels) previously observed to be learned in the early layers of ConvNets.  Brightness corresponds to pointwise function values. The majority of the plots show characteristics of oriented bandpass filters in two spatial dimensions, i.e., oscillating values along one direction, while remaining relatively constant in the orthogonal direction, even as there is an overall amplitude fall-off with distance from the center. The specific examples derive from the early layers of a ResNet-50 architecture \cite{he2016} trained on ImageNet \cite{russakovsky2015}.}
\end{figure}

\subsection{Related research}
\label{Sec:related}
Visualization of receptive field profiles (i.e., pointspread functions \cite{lim}) of the convolutional filters learned by contemporary ConvNets is a popular tool for providing insight into the image properties that are being represented by a network. A notable trend across these studies is that early layers appear to learn oriented bandpass filters in both two spatial dimensions, e.g., \cite{Zeiler2014, springenberg2015, yosinski2015, SHANG2016}, as well as three spatiotemporal dimensions, e.g., \cite{Feichtenhofer2018}. Indeed, earlier studies with architectures that also constrained their filters to be convolutional in nature, albeit using a Hebbian learning strategy \cite{mackay2003} rather than the currently dominant back-propagation approach \cite{bp}, also yielded filters that visualized as having oriented bandpass filter characteristics \cite{linsker1986}. Interestingly, biological vision systems also are known to show the presence of oriented bandpass filters at their earlier layers of processing in visual cortex; see \cite{hubel1962} for pioneering work along these lines and for more general review \cite{devalois}.

The presence of oriented bandpass filters in biological systems often has been attributed to their being well matched to the statistics of natural images \cite{field1987,olhausen1996,lewicki2009,simoncelli2001}, e.g., the dominance of oriented contours at multiple scales. Similar arguments have been made regarding why such filters are learned by ConvNets. Significantly, however, studies have shown that even when trained with images comprised of random noise patterns, convolutional architectures still learn oriented bandpass filters \cite{linsker1986}. These latter results suggest that the emergence of such filter tunings cannot be solely attributed to systems being driven to learn filters that were matched to their training data.  Similarly, recent work showed that randomly initialized networks serve well in image restoration problems \cite{Ulyanov2018}.

Some recent multilayer convolutional architectures have specified their earliest layers to have oriented bandpass characteristics, e.g., \cite{bruna2013,Jacobsen16,hadji2017}; indeed, some have specified such filters across all layers \cite{bruna2013,hadji2017}. These design decisions have been variously motivated in terms of being well matched to primitive image structure \cite{hadji2017} or providing useful building blocks for learning higher-order structures \cite{Jacobsen16} and capturing invariances \cite{bruna2013}. Other work has noted that purely 
mathematical considerations show that ConvNets are well suited to realizing filter designs 
for
capturing multiscale, windowed spectra \cite{bruna2016}; however, it did  not explicitly established the relationship to eigenfunctions of convolution nor offer an explanation for why deep-learning yields oriented bandpass filters when applied to ConvNets. It also did not provide empirical investigation of exactly what filter characteristics are learned at each convolutional layer of ConvNets.
\subsection{Contributions}

In the light of previous research, the present work appears to be the first to offer an explanation of why ConvNets learn oriented bandpass filters, independently of the input, by appeal to the inherent properties of their architectures in terms of their eigenfunctions. By definition, the convolutional layers of a ConvNet are governed by the properties of convolution. For present purposes, a key property is that the eigenfunctions of convolution are complex exponentials. Imposing locality on the eigenfunctions leads to oriented bandpass filters, which therefore are the appropriate filters to be learned by a ConvNet. Indeed, these theoretical considerations suggest that oriented bandpass filters should be learned at all layers of a ConvNet, not just at early layers. We provide empirical support for this observation by examining filters across all convolutional layers of three standard ConvNets (AlexNet \cite{Krizhevsky2012}, ResNet \cite{he2016} and VGG16 \cite{Simonyan2014}) and show that both numerically and visually they are well characterized as having learned oriented bandpass filters at all their convolutional layers. Our empirical study is distinct from earlier visualization efforts, which concentrate on the cumulative compositional results of filtering across layers that typically show emergence of complicated structures in the layerwise feature maps, while we focus on the complementary question of what primitive filter characteristics have been learned at each individual layer and offer both numerical as well as visualization analyses.

\section{Theory}
\label{sec:analysis}

This section details a novel explanation for why ConvNets learn oriented bandpass filters. The first two subsections largely review standard material regarding linear systems theory \cite{Oppenheim} and related topics \cite{kaiser2011,kusse2006}, but are necessary to motivate properly our explanation. The final subsection places the material in the 
context of ConvNets.
\subsection{Eigenfunctions of convolution}
\label{sec:eigenfunctions}

Let $\mathcal{L}$ be a linear operator on a function space. The set of eigenfunctions $\phi_n$ associated with this operator satisfy the condition \cite{kusse2006}
\begin{equation}
\mathcal{L} \phi_n = \lambda_n \phi_n.
\label{eq:eigenf}
\end{equation}
That is, the operator acts on the eigenfunctions simply via multiplication with a constant, $\lambda_n$, referred to as the eigenvalue. It sometimes also is useful to introduce a (positive definite) weighting function, $w$, which leads to the corresponding constraint
\begin{equation}
\mathcal{L} \phi_n = \lambda_n w \phi_n.
\label{eq:eigenfw}
\end{equation}
For cases where any function in the space can be expanded as a linear sum of the eigenfunctions, it is said that the collection of eigenfunctions form a complete set. Such a set provides a convenient and canonical spanning representation. 

Let $\mathbf{x} = (x_1, x_2, \ldots, x_n)^{\top}$, 
$\mathbf{a} = (a_1, a_2, \ldots, a_n)^{\top}$ and $\mathbf{u} = (u_1, u_2, \ldots, u_n)^{\top}$. For the space of convolutions, with the convolution of two functions, $f(\mathbf{x})$ and $h(\mathbf{x})$ defined as
\begin{equation}
f(\mathbf{x}) * h(\mathbf{x}) = \int_{- \infty}^{\infty} f(\mathbf {x} - \mathbf{a})h(\mathbf{a})\, d\mathbf{a}
\label{eq:convolution}
\end{equation}
it is well known that functions of the form $f(\mathbf{x}) = e^{i \mathbf{u}^{\top}\mathbf{x}}$ are eigenfunctions of convolution \cite{Oppenheim}, i.e.,
\begin{equation}
\int_{- \infty}^{\infty} e^{i \mathbf{u}^{\top}(\mathbf{x} - \mathbf{a})} h(\mathbf{a})\, d\mathbf{a} = e^{i \mathbf{u}^{\top}\mathbf{x}} \int_{- \infty}^{\infty} e^{-i \mathbf{u}^{\top}\mathbf{a}}h(\mathbf{a})\, d\mathbf{a}
\label{eq:eigenfunction}
\end{equation}
with the equality achieved via appealing to $ e^{i \mathbf{u}^{\top}(\mathbf {x} - \mathbf{a})} = e^{i \mathbf{u}^{\top}\mathbf{x}}  e^{-i \mathbf{u}^{\top}\mathbf{a}}$ and subsequently factoring $e^{i \mathbf{u}^{\top}\mathbf{x}}$ outside the integral as it is independent of $\mathbf{a}$. The integral on the right hand side of (\ref{eq:eigenfunction}),
\begin{equation}
\int_{- \infty}^{\infty} e^{-i \mathbf{u}^{\top}\mathbf{a}}h(\mathbf{a})\, d\mathbf{a},
\label{eq:mtf}
\end{equation}
is the eigenvalue, referred to as the modulation transfer function (MTF) in signal processing \cite{Oppenheim}. Noting that $ e^{i \mathbf{u}^{\top}\mathbf{x}} = \cos(\mathbf{u}^{\top}\mathbf{x)} + i \sin(\mathbf{u}^{\top}\mathbf{x}) $  leads to the standard interpretation of $\mathbf{u}$ in terms of frequency of the function (e.g., input signal).

Given the eigenfunctions of convolution are parameterized in terms of their frequencies, it is useful to appeal to the Fourier transform of function $f(\mathbf{x})$, where we use the form \cite{horn1986}
\begin{equation}
\mathcal{F}(\mathbf{u}) = \int_{- \infty}^{\infty} f(\mathbf{x}) e^{-i \mathbf{u}^{\top}\mathbf{x}}\, d\mathbf{x},
\label{eq:fourier}
\end{equation}
because any convolution can be represented in terms of how it operates via simple multiplication of the eigenvalues, (\ref{eq:mtf}), with the eigenfunctions, $ e^{i \mathbf{u}^{\top}\mathbf{x}}$, with $\mathbf{u}$ given by (\ref{eq:fourier}). Thus, this decomposition provides a canonical way to decompose $f(\mathbf{x})$ and explicate how a convolution operates on it.
\subsection{Imposing locality}
\label{sec:locality}

Understanding convolution purely in terms of its eigenfunctions and eigenvalues provides only a global representation of operations, as notions of signal locality, $\mathbf{x}$, are lost in the global transformation to the frequency domain, $\mathbf{u}$. This state of affairs often is unsatisfactory from a representational point of view because one wants to understand the structure of the signal (e.g., an image) on a more local basis (e.g., one wants to detect objects as well as their image coordinates). This limitation can be ameliorated by defining a windowed Fourier transform \cite{kaiser2011}, as follows \cite{jahne2000}.

Let $w(\mathbf{x})$ be a windowing function that is positive valued, symmetric and monotonically decreasing from its center so as to provide greatest emphasis at its center. A Windowed Fourier Transform (WFT) of $f(\mathbf{x})$ can then be defined as
\begin{equation}
\mathcal{F}(\mathbf{u}_{c},\mathbf{x};w) = \int_{- \infty}^{\infty} f(\mathbf{a}) w(\mathbf{a} - \mathbf{x}) e^{-i \mathbf{u}_{c}^{\top}\mathbf{a}}\, d\mathbf{a}.
\label{eq:windowedFourier}
\end{equation}
Making use of the symmetry constraint that we have enforced on the windowing function allows for $w(\mathbf{x}) = w(-\mathbf{x})$ so that the WFT, (\ref{eq:windowedFourier}), can be rewritten as
\begin{equation}
\mathcal{F}(\mathbf{u}_{c},\mathbf{x};w) = \int_{- \infty}^{\infty} f(\mathbf{a}) w(\mathbf{x} - \mathbf{a}) e^{i \mathbf{u}_{c}^{\top}(\mathbf{x} - \mathbf{a})} e^{-i \mathbf{u}_{c}^{\top} \mathbf{x}}\, d\mathbf{a},
\label{eq:windowedFourier2}
\end{equation}
which has the form of a convolution
\begin{equation}
f(\mathbf{x}) * \left(w(\mathbf{x}) e^{i \mathbf{u}_{c}^{\top} \mathbf{x}}\right)
\label{eq:psfWFT}
\end{equation}
with the inclusion of an additional phase component, $e^{-i \mathbf{u}_{c}^{\top} \mathbf{x}}$.

To provide additional insight into the impact the WFT convolution, (\ref{eq:psfWFT}), has on the function, $f(\mathbf{x})$, it is useful to examine the pointspread function, $w(\mathbf{x})e^{i \mathbf{u}_{c}^{\top} \mathbf{x}}$, in the frequency domain by taking its Fourier transform (\ref{eq:fourier}), i.e., calculate its MTF. We have
\begin{equation}
\int_{- \infty}^{\infty} w(\mathbf{x})  e^{i \mathbf{u}_{c}^{\top}\mathbf{x}} e^{-i \mathbf{u}^{\top}\mathbf{x}}\, d\mathbf{x},
\label{eq:mtfWFT}
\end{equation}
which via grouping by coefficients of $\mathbf{x}$ becomes
\begin{equation}
\int_{- \infty}^{\infty} w(\mathbf{x}) e^{-i (\mathbf{u} - \mathbf{u}_{c})^{\top}\mathbf{x}}\, d\mathbf{x}.
\label{eq:mtfWFT2}
\end{equation}
Examination of (\ref{eq:mtfWFT2}) reveals that it is exactly the Fourier transform of the window function, cf. (\ref{eq:fourier}), as shifted to the center frequencies, $\mathbf{u}_{c}$. Thus, operation of the WFT convolution, (\ref{eq:psfWFT}), on a function, $f(\mathbf{x})$, passes central frequency, $\mathbf{u}_{c}$, relatively unattenuated, while it suppresses those that are further away from the central frequency according to the shape of the window function, $w(\mathbf{x})$, i.e., it operates as a bandpass filter. Thus, convolution with a bank of such filters with varying central frequencies, $\mathbf{u}_{c}$,  has exactly the desired result of providing localized measures of the frequency content of the function $f(\mathbf{x})$. 

Returning to the pointspread function itself,  $w(\mathbf{x})e^{i \mathbf{u}_{c}^{\top} \mathbf{x}}$, and recalling that $ e^{i \mathbf{u}^{\top}\mathbf{x}} = \cos(\mathbf{u}^{\top}\mathbf{x)} + i \sin(\mathbf{u}_{c}^{\top}\mathbf{x}) $,  it is seen that in the signal domain, $\mathbf{x}$, the filter will oscillate along the direction of $\mathbf{u}_{c}$ while remaining relatively constant in the orthogonal direction, even as there is an overall amplitude fall-off with distance from the center according to the shape of $w(\mathbf{x})$, i.e., we have an oriented bandpass filter.

\begin{figure}[t] \label{fig:gabor}
	\begin{center}$
		\begin{array}{cc}
		\includegraphics[width=0.16\linewidth]{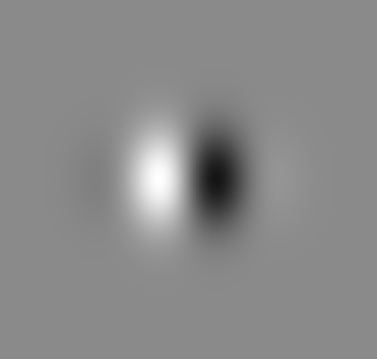} & 
		\includegraphics[width=0.15\linewidth]{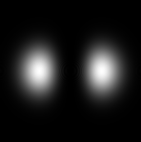}  
		\end{array}$
	\end{center}
	\caption{Visualization of an analytically defined oriented bandpass filter (\ref{eq:gabor}). The left panel shows the pointspread function corresponding to the odd symmetry ($\sin$) component, while the right panel shows its power in the frequency domain.  Brightness corresponds to pointwise function values. }
\end{figure}

As a specific example \cite{jahne2000}, taking $w(\mathbf{x})$ to be an n-dimensional Gaussian-like function, $g(\mathbf{x}; \sigma) =\kappa\, e^{- \|\mathbf{x} \|^{2}/\sigma^2}$, with $\sigma$ the standard deviation and $\kappa$ a scaling factor, yields an n-dimensional Gabor-like filter, 
\begin{equation}
g(\mathbf{x}; \sigma) e^{i \mathbf{u}_{c}^{\top}\mathbf{x}} = g(\mathbf{x}; \sigma) \left(\cos (\mathbf{u}_{c}^{\top}\mathbf{x})
+ i \sin (\mathbf{u}_{c}^{\top}\mathbf{x})\right),
\label{eq:gabor}
\end{equation}
which provides good joint localization of signal content in the signal and frequency domains \cite{gabor1946}. Indeed, visualization of these filters in two spatial dimensions (Figure 2) provides strikingly similar appearance to those presented in Figure 1, if in an idealized form. In particular, their pointspread functions oscillate according to a frequency $\| \mathbf{u}_{c} \|$ along the direction, $\frac{\mathbf{u}_{c}}{ \| \mathbf{u}_{c} \|}$, while remaining relatively constant in the orthogonal direction, even as there is an overall amplitude fall-off with distance from the center. In the frequency domain, they have peak power at $\mathbf{u}_{c}$ with a fall-off following a Gaussian-like shape with standard deviation, $1/\sigma$, that is the inverse of that used in specifying the window, $ w(\mathbf{x})$. These observations hold because we already have seen, (\ref{eq:mtfWFT2}), that the frequency domain representation of such a function is the Fourier transform of the window function, $w(\mathbf{x})$, shifted to the center frequencies, $\mathbf{u}_c$; furthermore, the Fourier transform of a function of the form $g(\mathbf{x}; \sigma)$ has a similar form, albeit with an inverse standard deviation \cite{bracewell}.

\subsection{Implications for ConvNets}
\label{subsec:implications}
Convolutions in ConvNets serve to filter the input signal to highlight its features according to the learned pointspread functions (convolutional kernels). Thus, convolution with the oriented filters shown in Figure 1  will serve to highlight aspects of an image that are correspondingly oriented and at corresponding scales. The question at hand is, ``Why did the ConvNet learn such filters?'' The previous parts of this section have reviewed the fact that complex exponentials of the form $e^{i \mathbf{u}^{\top}\mathbf{x}} = \cos(\mathbf{u}^{\top}\mathbf{x)} + i \sin(\mathbf{u}^{\top}\mathbf{x}) $ are the eigenfunctions of convolution. Thus, such frequency dependent functions similarly serve as the eigenfunctions of the convolutional operations in ConvNets. In particular, this result is a basic property of the convolutional nature of the architecture, independent of the input to the system. Thus, for any convolution in a ConvNet the frequency dependent eigenfunctions, $e^{i \mathbf{u}^{\top}\mathbf{x}}$, provide a systematic way to represent their input.

As with the general discussion of locality presented in Subsection~\ref{sec:locality}, for the specifics of ConvNets it also is of interest to be able to characterize and operate locally on a signal. At the level of convolution, such processing is realized via pointspread functions that operate as bandpass filters, (\ref{eq:psfWFT}). Like any practical system, ConvNets will not capture a continuous range of bandpass characteristics, as given by $\mathbf{u}_{c}$ and the sampling will be limited by the number of filters the designer allows at each layer, i.e., as a metaparameter of the system. Nevertheless, making use of these filters provides a systematic approach to representing the input signal.

Overall, the very convolutional nature of ConvNets inherently constrains and even defines the filters that they learn, independent of their input or training. In particular, learning bandpass filters provides a canonical way to represent and operate on their input, as these serve as the localized eigenfunctions of convolution. As a ConvNet is exposed to more and more training data, its representation is optimized by spanning as much of the data as it can. Within the realm of convolution, in which ConvNet conv layers are defined, oriented bandpass filters provide the solution. They arise as the locality constrained eigenfunctions of convolution and thereby have potential to provide a span of any input signal in a localized manner. Thus, ConvNets are optimized by learning exactly such filters. Notably, since this explanation for why ConvNets learn oriented bandpass filters is independent of training data, it can explain why such filters emerge even when the training data lacks such pattern structure, including training on random input signals, e.g., \cite{linsker1986}. Moreover, the explanation is independent of the learning algorithm as any algorithm driving its learned representation to span the space of input signals achieves its goal in the eigenfunctions of the convolutional architecture, i.e., oriented bandpass filters. Sec.~\ref{Sec:related} reviewed work showing that both back propagation and Hebbian learning yield oriented bandpass filters as their learned convolutional representations. 

Our analysis of oriented bandpass filters as the localized eigenfunctions of convolution is not specific to early ConvNet layers, but rather applies to any convolutional layer. The result thereby makes a theory-based prediction that oriented bandpass filters should be learned at all conv layers in a ConvNet. As reviewed in Sec.~\ref{Sec:related}, previous studies have demonstrated that filters learned at early ConvNet layers visualize as oriented bandpass filters; however, it appears that little attention in previous studies has examined the pointspread functions of learned convolutional filters deeper in a network. Instead, studies of learned filters deep in a ConvNet have focused on the cumulative effect of filtering across all layers upto and including a particular layer under consideration. In the following, we present a complementary study that empirically examines filters that have been learned at a given layer without reference to previous layers to see whether they appear as oriented bandpass filters.

\section{Empirical support}
\label{sec:empirical}
In this section, we present empirical support for the theory-based prediction that oriented bandpass filters should be learned at all layers in a ConvNet. We examine three standard ConvNets (AlexNet \cite{Krizhevsky2012}, ResNet50 \cite{he2016} and VGG16 \cite{Simonyan2014}) from both numerical and visualization perspectives. In all cases, we make use of publicly available implementations of the architectures \cite{alexnet,resnet,vgg16}. 

\begin{figure}[t] \label{fig:residuals}
	\begin{center}$
		\begin{array}{cc} \vspace{-10pt}
		\includegraphics[width=0.48\linewidth]{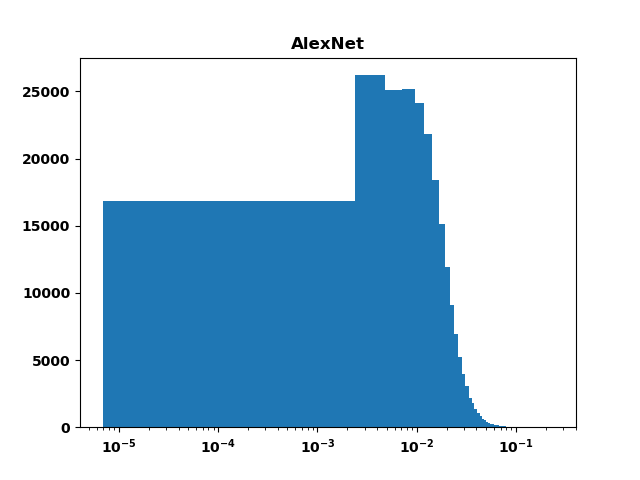} & 
		\includegraphics[width=0.48\linewidth]{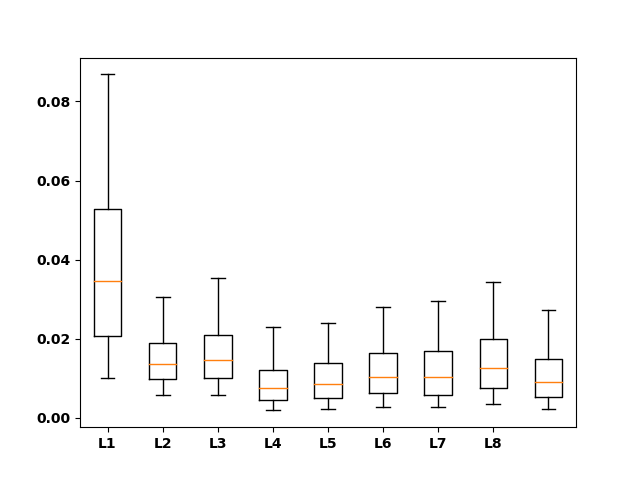}\\ \vspace{-10pt}
		\includegraphics[width=0.48\linewidth]{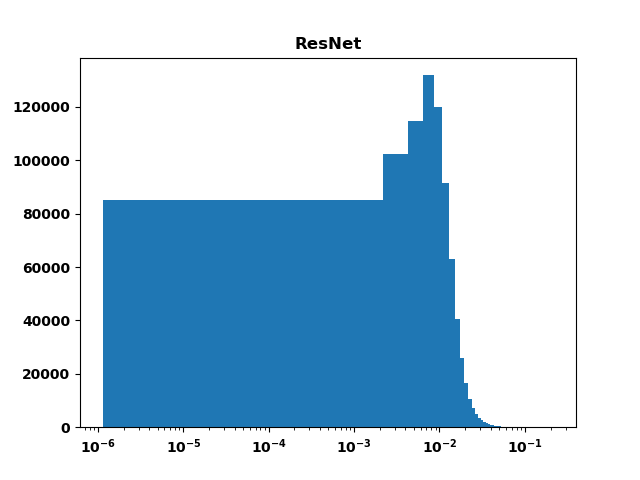} &
		\includegraphics[width=0.48\linewidth]{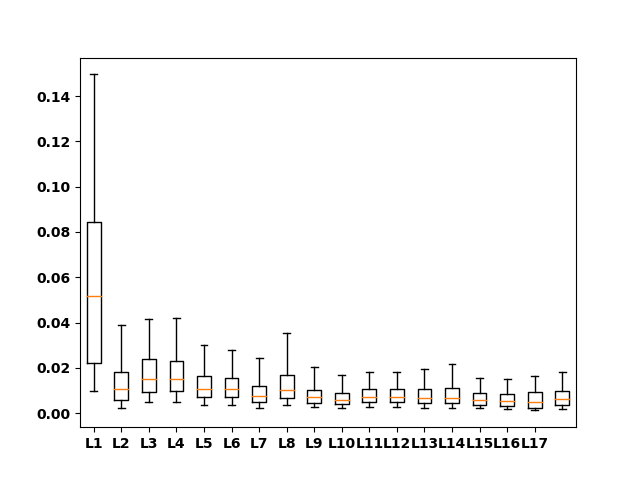}\\
		\includegraphics[width=0.48\linewidth]{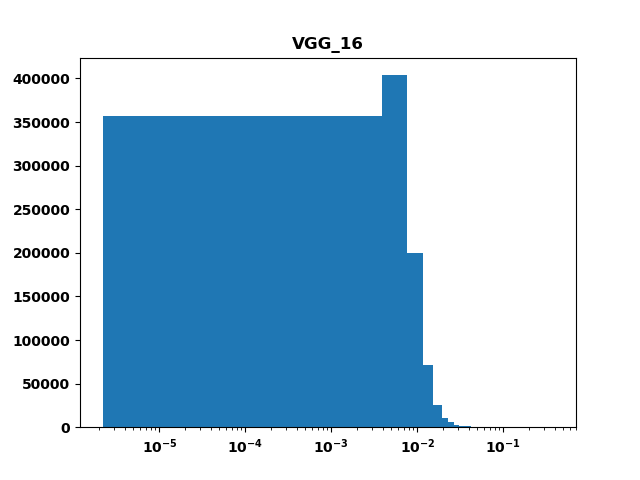} & \vspace{-10pt}
		\includegraphics[width=0.48\linewidth]{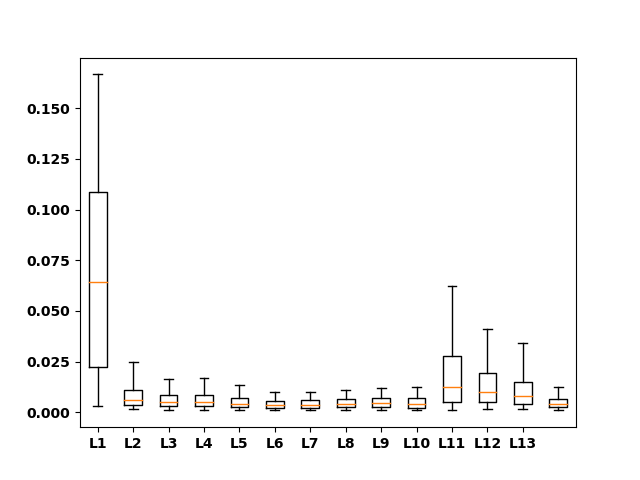}\\
		\end{array}$
	\end{center}
	\caption{RMS errors from fitting an oriented bandpass model to convolutional filters learned by AlexNet \cite{Krizhevsky2012} (top row), ResNet50 \cite{he2016}  (middle row) and VGG16 \cite{Simonyan2014} (bottom row). Left column shows histograms of residuals collapsed across all layers for each architecture: Residual values and count along the abscissa and ordinate, resp. Right column shows box plots \cite{tukey} of residuals by layer along abscissa, with the last plot for each architecture showing results across all layers: The box in each column encompasses the observed interquartile range, the horizontal line in each box indicates the median value, the whiskers above and below the box extend to the $95^{th}$ and $5^{th}$ percentiles, resp.} 
\end{figure}

\begin{figure}
	\label{fig:corruptions}
	\centering
	\includegraphics[width=0.45\linewidth]{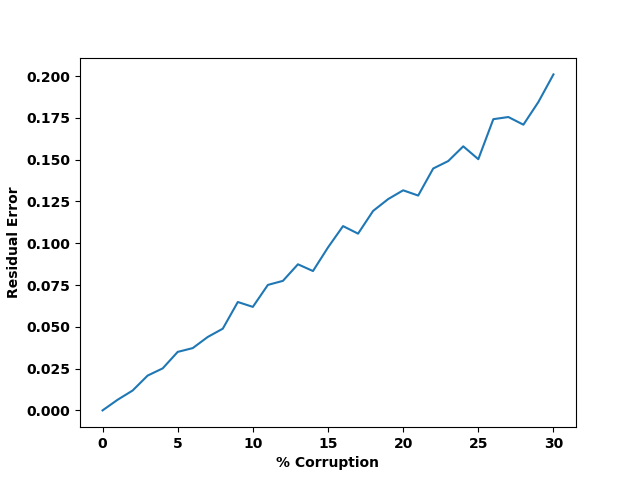}
	\caption{RMS error versus amount of corruption added to an analytically defined oriented bandpass filter to yield that error compared to the uncorrupted filter.} 
\end{figure}
\subsection{Numerical studies}
\label{subsec:numerical}
To study numerically whether ConvNets learn oriented bandpass filters, we perform a least-squares fit of the derived oriented bandpass filter, (\ref{eq:gabor}), to all learned convolutional filters at all layers of each model. While other oriented bandpass filter models can be used here
, the model considered, (\ref{eq:gabor}), is a natural choice in the present context, as it results from our theoretical analysis in Sec.~\ref{sec:analysis}. In particular, we fit the free parameters of a 2D instantiation of the model, (i.e., the center frequencies, $\mathbf{u}_c$, and the standard deviation, $\sigma$), to the learned pointspread values of every convolutional filter at every layer of AlexNet \cite{Krizhevsky2012}, ResNet50 \cite{he2016} and VGG16 \cite{Simonyan2014}. Finally, we take the root-mean-square (RMS) error residual 
between each individual fit of the model and the corresponding learned pointspread function as indicative of how well the learned filter is captured by an oriented bandpass filter, with smaller error indicative of a better fit. Results are plotted in Fig.~3. 

For all three architectures the histograms of residuals collapsed across layers shows that in all cases the fitting errors mostly lie below $0.1$, with the bulk of errors residing below $0.01$ and lower. These small residuals indicate generally good fits of the learned  filters to the oriented bandpass model.


A more detailed look can be had by considering box plots of errors by layer. Here, the results for AlexNet, for example, show that the median fitting error is under $0.04$ at layer one and subsequently decreased to under $0.02$ for all subsequent layers as well as for the aggregated fit across all layers. Moreover, $95\%$ of the data lies under $0.15$ at layer 1 and under $0.04$ at all other layers as well as the aggregate across layers. To place these numbers in perspective, we perturbed the otherwise analytically defined pointspread function, (\ref{eq:gabor}), with various amounts of noise and then compared to the same function without corruption to see how much corruption would yield various RMS errors. 
Results are summarized in Fig.~4. 
For example, we find that random noise within only $\approx$$6\%$ of the distribution of the uncorrupted filter values yields $0.04$ residual, which demonstrates that the discrepancy from the purely analytic form is very small, indicating that the observed fits are quite good.
Still, as indicated by the raw histograms, Fig.~3, some outliers are observed, where the errors approach $0.1$ and beyond. Visual inspection of these cases indicates that they arise when the learning process apparently has failed, as the learned filter has an essentially constant valued pointspread function (i.e., is flat) or else has no discernible structure.

Results for ResNet50 and VGG16 show similar patterns to those of AlexNet, with the main difference being that the distributions are shifted to larger values at their first layers; however, they subsequently conform to values similar to those of AlexNet thereafter. Also, it is seen that the layer one distribution for VGG16 is shifted upward relative to ResNet50. Perhaps this pattern arises because the layer one filter sizes are largest for AlexNet and smallest for VGG16, with ResNet50 lying in between, while at layer two and beyond VGG16 and ResNet50 have the same size filters and all three architectures have the same size filters at layer four and beyond. Interestingly, for all architectures it is seen that there is a marked decrease in the residual values between layers one and two. This result parallels observations made in earlier visualization studies that learned filters typically begin to show more strongly oriented structure after the first layer, e.g., \cite{linsker1986,springenberg2015}. Overall, these numerical experiments indicate that all three of the considered architectures learn oriented bandpass filters at all layers.

\subsection{Visualization studies}
\label{subsec:visualization}
\begin{figure}[t] \label{fig:visualization}
	\begin{center}$
		\begin{array}{ccccc}
		\mbox{\sf {\small Layer 1}} & & \mbox{\sf {\small Layer 2}}  & & \mbox{\sf {\small Layer 3}} \\
		\vcenteredhbox{
			\includegraphics[width=0.43\linewidth]{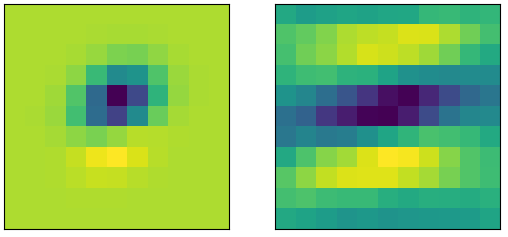}
		} 
		& &
		\vcenteredhbox{
			\includegraphics[width=0.215\linewidth]{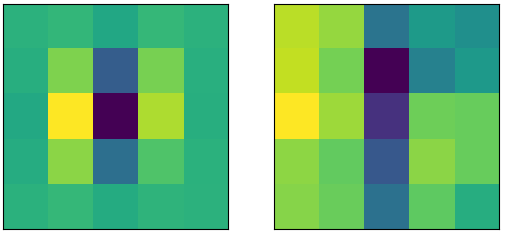} 
		}
		& &
		\vcenteredhbox{
			\includegraphics[width=0.215\linewidth]{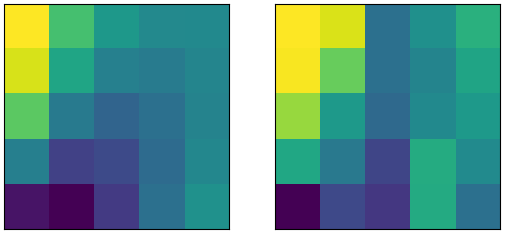} 
		}
		\end{array}$
	\end{center}
	\vspace{0pt}
	\begin{center}$
		\begin{array}{ccccccccc}
		\mbox{\sf {\small Layer 4}} & & \mbox{\sf {\small Layer 5}} & & \mbox{\sf {\small Layer 6}} & & \mbox{\sf {\small Layer 7}} & & \mbox{\sf {\small Layer 8}}
		\\
		\includegraphics[width=0.12\linewidth]{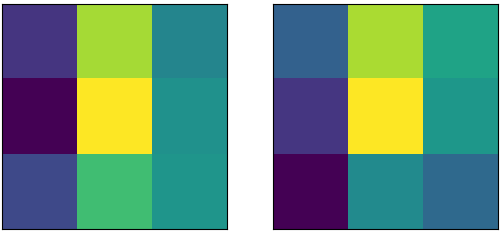} & &
		\includegraphics[width=0.12\linewidth]{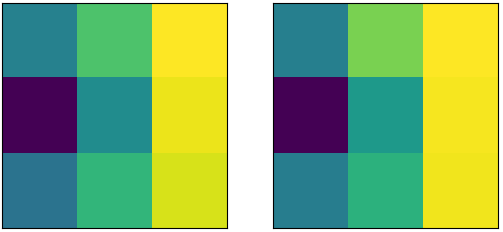} & &
		\includegraphics[width=0.12\linewidth]{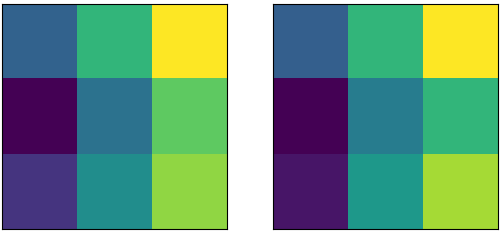} & &
		\includegraphics[width=0.12\linewidth]{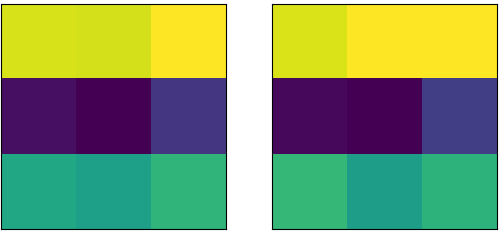} & &
		\includegraphics[width=0.12\linewidth]{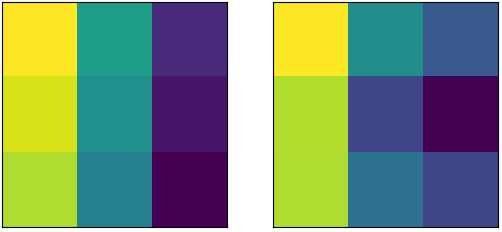} 
		\end{array}$
	\end{center}
	\caption{Visualization of learned pointspread functions for AlexNet \cite{Krizhevsky2012}
		. Shown are visualizations of representative functions that have median residual values at each layer. For each layer, the plot on the left shows the learned function and the plot on the right shows the correspondingly fit function. Brightness corresponds to pointwise function values.}
\end{figure}

For visualization studies, we plot the pointspread functions for a selection of the learned filters at each layer and display them as images. In the interest of space, here we focus on AlexNet; although, the visualization results for the other architectures similarly are supportive of oriented bandpass filters, as would be expected from the numerical results of Sec.~\ref{subsec:numerical}. In particular, Fig.~5 shows plots of learned pointspread functions from all layers of AlexNet. The shown pointspread functions are representative of the median residual value for each layer, and they are paired with a visualization of the corresponding fit to the oriented bandpass model, (\ref{eq:gabor}).
Inspection of these plots show that in all cases oriented structure is visible in the learned pointspread functions: In particular, proceeding from layers 1 to 8 orientations are manifest approximately along slight diagonal upper right to lower left, vertical, diagonal upper right to lower left, vertical, vertical, diagonal upper left to lower right, horizontal and vertical, resp. Moreover, the learned and fit pointspread functions are qualitatively very similar. Moreover, the visualizations suggest improved fits between the learned and fit models as layer increases, similar to what is seen in the numerical results of Sec.~\ref{subsec:numerical}. Overall, these visualization results corroborate the numerical results indicating that oriented bandpass filters are indeed learned at all layers of the considered ConvNets.

\section{Summary}
\label{sec:summary}
Previous studies have demonstrated that learned filters at the early layers of convolutional networks visualize as oriented bandpass filters. This phenomenon typically is explained via appeal to natural image statistics, i.e., natural images are dominated by oriented contours manifest across a variety of scales and oriented bandpass filters are well matched to such structure. We have offered an alternative explanation in terms of the structure of convolutional networks themselves. Given that their convolutional layers necessarily operate within the space of convolutions, learning oriented bandpass filters provides the system with the potential to span possible input, even while preserving a notion of locality in the signal domain. Notably, our work is applicable to not just early ConvNet layers, but to all conv layers in such networks. We have provided empirical support for this claim, showing that oriented bandpass filters are indeed learned at all layers of three standard ConvNets. These results not only provide new insights into the operations and representations learned by ConvNets, but also suggest interesting future research. In particular, our work motivates investigation of novel architectures that explicitly constrain their convolutional filters to be oriented bandpass in a learning-based framework. In such a framework, it would not be necessary for the training process to learn the numerical values for each and every individual filter value (i.e., each filter tap), but rather would merely need to learn a much smaller number of parameters, e.g., the values of the center frequency, $\mathbf{u}_c$, and the standard deviation, $\sigma$, associated with the Gabor-like filter derived above, (\ref{eq:gabor}), or some other suitably parameterized filter. Such a constrained  learning approach would require a much less intensive training procedure (e.g. involving far less data)  compared to learning values for all  individual filter taps,  owing to  the drastically reduced number of parameters that need to be estimated, even while being able to tune to the specifics of the task that is being optimized.

\bibliography{mybibtex}
\bibliographystyle{iclr2021_conferenceMy}
\end{document}